\documentclass[runningheads]{llncs}

\usepackage[final,year=2024,ID=4689]{eccv}

\usepackage{eccvabbrv}

\usepackage[dvipsnames]{xcolor}
\usepackage{epsfig}
\usepackage{graphicx}
\usepackage{float}
\usepackage{amsmath}
\usepackage{pifont}
\usepackage{adjustbox}
\usepackage{amssymb}
\usepackage{multirow}
\usepackage{booktabs}
\usepackage{pgfplots}
\usepackage{colortbl}
\usepackage[accsupp]{axessibility}
\usepackage{hhline}
\usepackage[skip=1ex,font=small,labelsep=period]{caption}
\usepackage{enumitem}
\usepackage{tabularx}
\usepackage{mathtools}
\pgfplotsset{compat=1.18} 
\tikzset{cross/.style={cross out, draw=black, minimum size=2*(#1-\pgflinewidth), inner sep=0pt, outer sep=0pt},
cross/.default={5pt}}

\usetikzlibrary{arrows.meta,shapes.arrows}

\usepackage[pagebackref,breaklinks,colorlinks,citecolor=eccvblue]{hyperref}

\usepackage{orcidlink}

\begin{document}

\newcommand{\enric}[1]{{\color{orange}\textbf{EC: #1}}}
\newcommand{\andreiz}[1]{{\color{purple}\textbf{AZ: #1}}}
\newcommand{\thiemo}[1]{{\color{blue}\textbf{TA: #1}}}

\newcommand{\Model}{VLOGGER\xspace}
\newcommand{\Dataset}{MENTOR\xspace}

\newcommand{\mA}{\mathcal{A}}
\newcommand{\mF}{\mathcal{F}}
\newcommand{\mI}{\mathcal{I}}
\newcommand{\mK}{\mathcal{K}}
\newcommand{\mP}{\mathcal{P}}
\newcommand{\mR}{\mathcal{R}}
\newcommand{\mcU}{\mathcal{U}}
\newcommand{\mZ}{\mathcal{Z}}
\newcommand{\suppmat}{Sup.\ Mat\onedot}
\newcommand{\suppvid}{Sup.\ Video}

\definecolor{markgreen}{HTML}{27ae60}
\definecolor{markred}{HTML}{e74c3c}
\newcommand{\cmark}{\textcolor{markgreen}{\ding{51}}}%
\newcommand{\xmark}{\textcolor{markred}{\ding{55}}}%
\newcolumntype{R}[2]{%
    >{\adjustbox{angle=#1,lap=\width-(#2)}\bgroup}%
    l%
    <{\egroup}%
}
\newcommand*\rot{\multicolumn{1}{R{12}{1em}}}

\renewcommand{\vec}[1]{\boldsymbol{#1}}
\newcommand{\mat}[1]{\mathbf{#1}}
\newcommand{\set}[1]{\mathcal{#1}}

\newcommand{\real}[0]{\mathbb{R}}
\newcommand{\imageset}[0]{\set{I}}
\newcommand{\image}[0]{\mat{I}}

\newcommand{\blendweight}[0]{w}
\newcommand{\blendweights}[0]{\mat{W}}

\newcommand\customparagraph[1]{\vspace{0.7em}\noindent\textbf{#1}}
\renewcommand{\paragraph}[1]{\noindent\textbf{#1}\:}

\definecolor{butter1}{rgb}{0.988,0.914,0.310}
\definecolor{butter2}{rgb}{0.929,0.831,0.000}
\definecolor{butter3}{rgb}{0.769,0.627,0.000}
\definecolor{orange1}{rgb}{0.988,0.686,0.243}
\definecolor{orange2}{rgb}{0.961,0.475,0.000}
\definecolor{orange3}{rgb}{0.808,0.361,0.000}
\definecolor{chocolate1}{rgb}{0.914,0.725,0.431}
\definecolor{chocolate2}{rgb}{0.757,0.490,0.067}
\definecolor{chocolate3}{rgb}{0.561,0.349,0.008}
\definecolor{chameleon1}{rgb}{0.541,0.886,0.204}
\definecolor{chameleon2}{rgb}{0.451,0.824,0.086}
\definecolor{chameleon3}{rgb}{0.306,0.604,0.024}
\definecolor{skyblue1}{rgb}{0.447,0.624,0.812}
\definecolor{skyblue2}{rgb}{0.204,0.396,0.643}
\definecolor{skyblue3}{rgb}{0.125,0.290,0.529}
\definecolor{plum1}{rgb}{0.678,0.498,0.659}
\definecolor{plum2}{rgb}{0.459,0.314,0.482}
\definecolor{plum3}{rgb}{0.361,0.208,0.400}
\definecolor{scarletred1}{rgb}{0.937,0.161,0.161}
\definecolor{scarletred2}{rgb}{0.800,0.000,0.000}
\definecolor{scarletred3}{rgb}{0.643,0.000,0.000}
\definecolor{aluminium1}{rgb}{0.933,0.933,0.925}
\definecolor{aluminium2}{rgb}{0.827,0.843,0.812}
\definecolor{aluminium3}{rgb}{0.729,0.741,0.714}
\definecolor{aluminium4}{rgb}{0.533,0.541,0.522}
\definecolor{aluminium5}{rgb}{0.333,0.341,0.325}
\definecolor{aluminium6}{rgb}{0.180,0.204,0.212}

\definecolor{tabfirst}{rgb}{1, 0.7, 0.7} %
\definecolor{tabsecond}{rgb}{1, 0.85, 0.7} %
\definecolor{tabthird}{rgb}{1, 1, 0.7} %

\newcommand{\ba}{\boldsymbol{a}}
\newcommand{\bb}{\boldsymbol{b}}
\newcommand{\bc}{\boldsymbol{c}}
\newcommand{\bd}{\boldsymbol{d}}
\newcommand{\be}{\boldsymbol{e}}
\newcommand{\bff}{\boldsymbol{f}}
\newcommand{\bg}{\boldsymbol{g}}
\newcommand{\bh}{\boldsymbol{h}}
\newcommand{\bi}{\boldsymbol{i}}
\newcommand{\bj}{\boldsymbol{j}}
\newcommand{\bl}{\boldsymbol{l}}
\newcommand{\bn}{\boldsymbol{n}}
\newcommand{\bo}{\boldsymbol{o}}
\newcommand{\bp}{\boldsymbol{p}}
\newcommand{\bq}{\boldsymbol{q}}
\newcommand{\br}{\boldsymbol{r}}
\newcommand{\bs}{\boldsymbol{s}}
\newcommand{\bts}{\tilde{\boldsymbol{s}}}
\newcommand{\bt}{\boldsymbol{t}}
\newcommand{\bu}{\boldsymbol{u}}
\newcommand{\btu}{\tilde{\boldsymbol{u}}}
\newcommand{\bv}{\boldsymbol{v}}
\newcommand{\btv}{\tilde{\boldsymbol{v}}}
\newcommand{\bbv}{\bar{\boldsymbol{v}}}
\newcommand{\bw}{\boldsymbol{w}}
\newcommand{\bx}{\boldsymbol{x}}
\newcommand{\btx}{\tilde{\boldsymbol{x}}}
\newcommand{\by}{\boldsymbol{y}}
\newcommand{\bty}{\tilde{\boldsymbol{y}}}
\newcommand{\bz}{\boldsymbol{z}}

\newcommand{\bA}{\mathbf{A}}
\newcommand{\bB}{\mathbf{B}}
\newcommand{\bC}{\mathbf{C}}
\newcommand{\bD}{\mathbf{D}}
\newcommand{\bE}{\mathbf{E}}
\newcommand{\bF}{\mathbf{F}}
\newcommand{\btF}{\tilde{\mathbf{F}}}
\newcommand{\bG}{\mathbf{G}}
\newcommand{\bH}{\mathbf{H}}
\newcommand{\bI}{\mathbf{I}}
\newcommand{\bJ}{\mathbf{J}}
\newcommand{\bK}{\mathbf{K}}
\newcommand{\bL}{\mathbf{L}}
\newcommand{\bM}{\mathbf{M}}
\newcommand{\bN}{\mathbf{N}}
\newcommand{\bO}{\mathbf{O}}
\newcommand{\bP}{\mathbf{P}}
\newcommand{\btP}{\tilde{\mathbf{P}}}
\newcommand{\bQ}{\mathbf{Q}}
\newcommand{\bR}{\mathbf{R}}
\newcommand{\btR}{\tilde{\mathbf{R}}}
\newcommand{\btS}{\tilde{\mathbf{S}}}
\newcommand{\bS}{\mathbf{S}}
\newcommand{\bT}{\mathbf{T}}
\newcommand{\btT}{\tilde{\mathbf{T}}}
\newcommand{\bU}{\mathbf{U}}
\newcommand{\bV}{\mathbf{V}}
\newcommand{\btV}{\tilde{\mathbf{V}}}
\newcommand{\bW}{\mathbf{W}}
\newcommand{\bX}{\mathbf{X}}
\newcommand{\btX}{\tilde{\mathbf{X}}}
\newcommand{\bY}{\mathbf{Y}}
\newcommand{\btY}{\tilde{\mathbf{Y}}}
\newcommand{\bZ}{\mathbf{Z}}

\newcommand{\1}{\bmath{1}}
\newcommand{\0}{\bmath{0}}

\newcommand{\bhA}{\hat{\mathbf{A}}}
\newcommand{\bhc}{\hat{\mathbf{c}}}
\newcommand{\bhC}{\hat{\mathbf{C}}}
\newcommand{\bhd}{\hat{\mathbf{d}}}
\newcommand{\bhF}{\hat{\mathbf{F}}}
\newcommand{\bhG}{\hat{\mathbf{G}}}
\newcommand{\bhH}{\hat{\mathbf{H}}}
\newcommand{\bhJ}{\hat{\mathbf{J}}}
\newcommand{\bhj}{\hat{\mathbf{j}}}
\newcommand{\bhK}{\hat{\mathbf{K}}}
\newcommand{\bhM}{\hat{\mathbf{M}}}
\newcommand{\bhw}{\hat{\mathbf{w}}}
\newcommand{\bhx}{\hat{\mathbf{x}}}
\newcommand{\bhX}{\hat{\mathbf{X}}}
\newcommand{\bhY}{\hat{\mathbf{Y}}}

\newcommand{\Avvo}{\mathcal{A}}
\newcommand{\Tvvo}{\mathcal{T}}
\newcommand{\balpha}{\boldsymbol{\alpha}}
\newcommand{\bdelta}{\boldsymbol{\delta}}
\newcommand{\bDelta}{\boldsymbol{\Delta}}
\newcommand{\blambda}{\boldsymbol{\lambda}}
\newcommand{\bLambda}{\boldsymbol{\Lambda}}
\newcommand{\bmu}{\boldsymbol{\mu}}
\newcommand{\bgamma}{\boldsymbol{\gamma}}
\newcommand{\bGamma}{\boldsymbol{\Gamma}}
\newcommand{\bsigma}{\boldsymbol{\sigma}}
\newcommand{\bSigma}{\boldsymbol{\Sigma}}
\newcommand{\btheta}{\boldsymbol{\theta}}
\newcommand{\bTheta}{\boldsymbol{\Theta}}
\newcommand{\brho}{\boldsymbol{\rho}}
\newcommand{\bphi}{\boldsymbol{\phi}}
\newcommand{\bPhi}{\boldsymbol{\Phi}}
\newcommand{\bpsi}{\boldsymbol{\psi}}
\newcommand{\bPsi}{\boldsymbol{\Psi}}
\newcommand{\bxi}{\boldsymbol{\xi}}
\newcommand{\bUpsilon}{\boldsymbol{\Upsilon}}
\newcommand{\bomega}{\boldsymbol{\omega}}
\newcommand{\bOmega}{\boldsymbol{\Omega}}

\newcommand\blfootnote[1]{%
  \begingroup
  \renewcommand\thefootnote{}\footnote{#1}%
  \addtocounter{footnote}{-1}%
  \endgroup
}

\newcommand{\audio}{\vec{a}}
\newcommand{\samplerate}{S}
\newcommand{\videolength}{N}
\newcommand{\width}{W}
\newcommand{\height}{H}
\newcommand{\identity}{I}
\newcommand{\numjoints}{J}

\newcommand{\video}{\boldsymbol{\bV}}
\newcommand{\controls}{\mathbf{C}}

\newcommand{\motionnetwork}{f}
\newcommand{\videonetwork}{f}

\newcommand{\inputdiffusion}[0]{\mathbf{Z}}
\newcommand{\expression}[0]{\mathbf{\theta}^{e}}
\newcommand{\headpose}[0]{\mathbf{\theta}^{h}}
\newcommand{\bodypose}[0]{\mathbf{\theta}^{b}}
\newcommand{\shape}[0]{\vec{\beta}}
\newcommand{\colorcode}[0]{\vec{c}}
\newcommand{\trans}[0]{\vec{t}}
\newcommand{\joints}[0]{\mat{J}}
\newcommand{\transformations}[0]{\mat{T}}
\newcommand{\colormesh}[0]{\mat{C}}

\newcommand{\loss}{\mathcal{L}}

\title{
\vspace{-0.3cm}
\Model: Multimodal Diffusion for Embodied Avatar Synthesis
\vspace{-0.35cm}
} 

\titlerunning{\Model: Multimodal Diffusion for Embodied Avatar Synthesis}

\author{
Enric Corona \qquad Andrei Zanfir \qquad Eduard Gabriel Bazavan \\ Nikos Kolotouros \qquad  Thiemo Alldieck \qquad Cristian Sminchisescu \vspace{2mm}
}

\authorrunning{E.~Corona et al.}

\institute{
\normalsize{Google Research}
\\
\url{https://enriccorona.github.io/vlogger/}}

\maketitle

\begin{figure*}[h]
\includegraphics[width=1.0\linewidth, trim={0cm, 0cm 0cm 0cm}, clip=true]{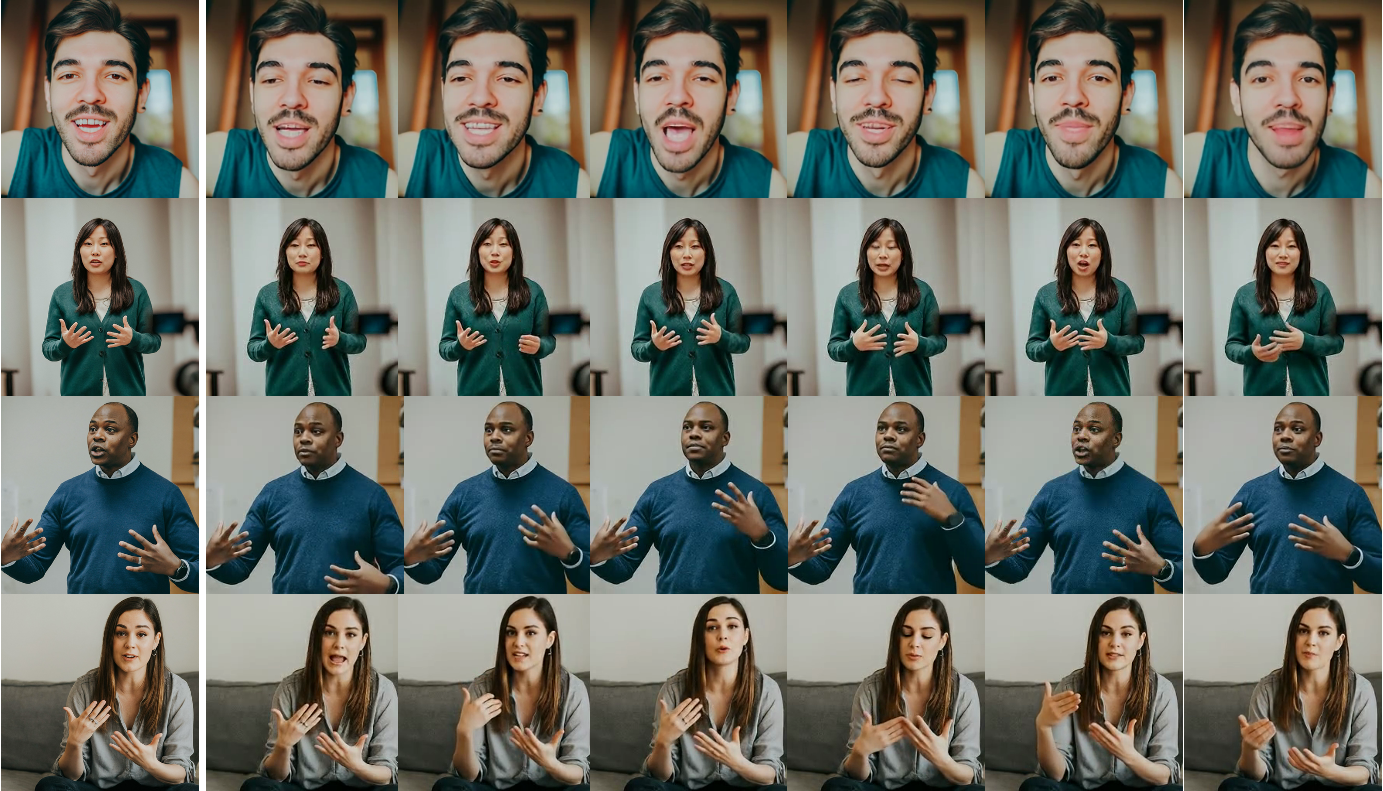}\\
\begin{tikzpicture}
  \hspace{-0.14cm}
  \draw node (0,0) {\footnotesize{Input Image}};
  \draw[-{Stealth[black]},line width=1.3pt] (1.2,0)   -- (11.3,0) 
  node [midway,fill=white] {\footnotesize{Generated Video}};
\end{tikzpicture}
\caption{
\Model is a novel framework to synthesize humans from audio. Given a single input image like the ones shown on the first column, and a sample audio input, our method generates photorealistic and temporally coherent videos of the person talking and vividly moving. As seen on the synthesized images in the right columns, we generate head motion, gaze, blinking, lip movement and unlike previous methods, upper-body and hand gestures, thus taking audio-driven synthesis one step further.
\label{fig:teaser}
}
\end{figure*}

\begin{abstract}
We propose \Model, a method for audio-driven human video generation from a single input image of a person, which builds on the success of recent generative diffusion models. Our method consists of 1) a stochastic human-to-3d-motion diffusion model, and 2) a novel diffusion-based architecture that augments text-to-image models with both spatial and temporal controls. This supports the generation of high quality video of variable length, easily controllable through high-level representations of human faces and bodies. In contrast to previous work, our method does not require training for each person, does not rely on face detection and cropping, generates the complete image (not just the face or the lips), and considers a broad spectrum of scenarios (\eg visible torso or diverse subject identities) that are critical to correctly synthesize humans who communicate. 
We also curate \Dataset, a new and diverse dataset with 3d pose and expression annotations, one order of magnitude larger than previous ones (800,000 identities) and with dynamic gestures, on which we train and ablate
our main technical contributions.

\Model outperforms state-of-the-art methods in three public benchmarks, considering image quality, identity preservation and temporal consistency while also generating upper-body gestures. 
We analyze the performance of \Model with respect to multiple diversity metrics, showing that our architectural choices and the use of \Dataset benefit training a fair and unbiased model at scale. Finally we show applications in video editing and personalization.
\end{abstract}

\section{Introduction}
\label{sec:intro}

We present \Model, a method to automatically generate a video of a talking and moving person, based on text or audio, and given only a single image of that person.
Industries like content creation, entertainment, or gaming all have high demand for human synthesis.
Yet, the creation of realistic videos of humans is still complex and ripe with artifacts. This requires significant manual intervention for realistic results. %
Full automation, however, would not only ease creative processes, but also enable entirely new use cases, such as enhanced online communication, education, or personalized virtual assistants, to name a few.
The latter is especially relevant, given the recent success of chat agents \cite{gpt4,manyika2023overview}. Research has shown that such solutions are not perceived as natural enough to develop empathy \cite{zhou2023talking} and several authors \cite{kyrlitsias2022social} argue that anthropomorphism and behavioral realism (\eg gaze, facial expressions, whole-body movements, \etc) are critical in creating a social presence and in eliciting empathetic responses from the user. Such features would result in the wide adoption of agents \cite{moussawi2021perceptions}, in areas like customer service \cite{adam2021ai, pizzi2023chatbot}, telemedicine \cite{seitz2022can}, education \cite{seeger2021texting}, or human-robot interaction \cite{roesler2021meta}. 
It is precisely automation and behavioral realism that what we aim for in this work:
\Model is a multi-modal interface to an \textit{embodied conversational agent} \cite{wahde2022conversational}, equipped with an audio and animated visual representation, featuring complex facial expressions and increasing level of body motion, designed to support natural conversations with a human user.
\Model can be used as a stand-alone solution for presentations, education, narration, low-bandwidth online communication, and as an interface for text-only  HCI \cite{zhou2023comprehensive, bard2023}. %
In this paper, we additionally illustrate its potential in video editing tasks.

Multimodal, photorealistic human synthesis, is  complex due to challenges like data acquisition, enacting facial expressions in a natural way, expression to audio synchronization, occlusion, or representing full-body movements ---  especially given a single input image.
Many attempts focused exclusively on lip sync \cite{wav2lip, wu2023speech2lip, wang2023seeing}, by editing the mouth region of a driving video. Recently, \cite{thpad, sadtalker} relied on  extensive advances in face reenactment \cite{PECHead, zhang2019one,nirkin2019fsgan,thies2016face2face,hsu2022dual,yang2022face2face,bounareli2023hyperreenact} to generate talking head videos from a single image by predicting face motion from audio. 
Temporal consistency is usually achieved with a per-frame image generation network by relying on a smooth guiding motion from face keypoints. 
However, this might cause blurriness and does not ensure temporal coherency in areas more distant from the face.
Consequently, most methods require detecting and cropping the head, whenever a significant part of the body is visible.
In this paper, we argue that communication is more than ``just'' audio combined with lip and face motion -- humans communicate using their body via gestures, gaze, blinks, or pose. 
MODA~\cite{moda} recently started exploring the animation of both face and body, however in limited scenarios, and without generalization to new identities.
In contrast, we aim for a \emph{general, person agnostic synthesis solution}, focusing on realism and diversity in motion, including both head and hand gestures. 
Our objective is to bridge the gap between recent video synthesis efforts~\cite{kondratyuk2023videopoet,blattmann2023stable,bar2024lumiere,sora}, which can generate dynamic videos with no control over identity or pose, and controllable image generation methods~\cite{PECHead,bounareli2023hyperreenact,dreambooth}.

Towards that goal, we propose a two-step approach where first a generative diffusion-based network predicts body motion and facial expressions according to an input audio signal. This stochastic approach is necessary to model the nuanced (one-to-many) mapping between speech and pose, gaze, and expression. Second, we propose and ablate a novel architecture based on recent image diffusion models, which provides control in the temporal and spatial domains. 
By additionally relying on generative human priors, acquired during pre-training, we show how this combined architecture improves the capacity of image diffusion models, which often struggle to generate consistent human images (\eg eyes).
VLOGGER consists of a base model followed by a super-resolution diffusion model to obtain high quality videos. 
We condition the video generation process on 2d controls that represent the full body, including facial expressions as in previous work, but also body and hands. To generate videos of arbitrary length, we follow a temporal outpainting approach to condition new video clips based on previous frames. 
Finally, the flexibility of VLOGGER enables editing particular parts of an input video, like lips or the face region.

For robustness and generalisation, we curate a large-scale dataset featuring a much larger diversity than previously available data, in terms of skin tone, body pose, viewpoint, speech and body visibility. 
In contrast to previous attempts, the dataset also contains videos with dynamic hand gestures, which are important in learning the complexity of human communication.
VLOGGER outperforms previous work across different diversity metrics, and obtains state-of-the-art image quality and diversity results on the previous HDTF~\cite{hdtf} and TalkingHead-1KH~\cite{talkinghead} datasets. 
Moreover, our method considers a larger range of scenarios than baselines, by generating high resolution video of head and upper-body motion, and by featuring considerably diverse facial expressions and gestures. 
Finally, in the experimental section we explore downstream applications, to demonstrate VLOGGER's flexibility and capacity to adapt to different scenarios. For instance, VLOGGER can be used for video editing by inpainting selected regions of each frame, such as the lips or the face, as well as for personalization.

To summarize, the main contributions are: 1) VLOGGER is the first approach to generate talking and moving humans given speech inputs; (2) leveraging a diverse, curated dataset, called MENTOR, which is one order of magnitude larger than existing ones, for training and testing our models; (3) A large ablation study that validates the proposed methodology on controlled video generation, comparing against existing diffusion-based solutions and showing the benefits of the proposed 2d body controls; (4) VLOGGER outperforms previous SOTA in large quantitative comparisons on three public benchmarks; (5) Diversity analysis where VLOGGER shows low bias and outperforms baselines on different perceived human attributes; (6) Applications of VLOGGER to video editing 
and an analysis of its stochasticity.

\begin{table}[t]
\centering
\setlength{\tabcolsep}{17pt}
\resizebox{\linewidth}{!}{%
    \begin{tabular}{ccccccc|l}
    \rot{Audio Control} & \rot{Face Control} & \rot{Body Control} & \rot{Stochastic} & \rot{Photorealistic} & \rot{Generalizes to new subjects} & \rot{Can edit videos} & \\ %
    \hline
    \xmark & \cmark & \xmark & \xmark & \cmark  & \cmark & \xmark  & Face Reenactment \cite{thies2016face2face,talkinghead}
    \\
    \cmark & \cmark & \xmark & \cmark & \xmark  & \cmark & \xmark  & Audio-to-Motion \cite{imitator,faceformer} \\
    \cmark & \xmark & \xmark & \xmark & \cmark  & \cmark & \cmark  & Lip Sync \cite{wav2lip,stylesync}
    \\
    \cmark & \cmark & \xmark & \cmark & \cmark & \cmark & \xmark  & SadTalker \cite{sadtalker} \\
    \cmark & \cmark & \xmark & \xmark & \cmark & \cmark & \xmark  & Styletalk \cite{styletalk} \\
    \hline
    \cmark & \cmark & \cmark & \cmark & \cmark & \cmark & \cmark & \textbf{\Model~(Ours)} \\
\end{tabular}}
\caption{
\textbf{Key properties of \Model compared to related  work}. Face Reenactment \cite{PECHead, zhang2019one,nirkin2019fsgan,thies2016face2face,hsu2022dual,yang2022face2face,bounareli2023hyperreenact} generally does not consider driving using audio or text. 
Works on audio-to-motion~\cite{imitator,facediffuser,faceformer,codetalker,voca,meshtalk,talkshow} shares components by encoding audio into 3d face motion, however lack photorealism. Lip sync~\cite{stylesync,wav2lip} consider input videos of different subjects, but only model mouth motion. Given their generalisation capacity, SadTalker \cite{sadtalker} and Styletalk~\cite{styletalk} are the closest to us, but require cropped images of faces, lack body control, and cannot edit videos.
}
\label{tab:positioning}
\end{table}

\section{Related Work}
\label{sec:relatedwork}

\paragraph{Audio-Driven Talking Face Generation.} There has been a significant amount of work in talking face generation, which can be categorized according to the driving inputs, intermediate representations and output formats. We provide an overview and comparison against our work in \cref{tab:positioning}.
There exists a body of work in animation of 3D morphable faces~\cite{imitator,facediffuser,faceformer,codetalker,voca,meshtalk} or full body~\cite{talkshow} models based on audio segments. These efforts can generate diverse 3d talking heads in the form of temporally coherent pose and shape parameters of various statistical head or body models~\cite{sphear2023, ghum2020, smplbogo2016, flameli2017learning, baselpaysan20093d}. We consider a similar network to guide the generated motion but, in this paper, we instead aim to generate photorealistic talking humans with diversity in expression and head or body motion, that are coherent with an image of a target subject. We consider challenges such as temporal consistency, subject diversity, hair, gaze, and detail in output videos.

In the image domain, incipient works have focused on the task of mouth editing \cite{chung2017said, jamaludin2019you, chen2019hierarchical, vougioukas2020realistic, wav2lip, hdtf}, such as only predicting the lip motion, synchronized with the input audio. Follow up works added extended features such as head motion, gaze and blinking 
\cite{ren2021pirenderer, zhou2021pose, lu2021live, glowguidedzhang2021, audio2headWang2021, eammSiggraph2022}, using intermediate 2d, 3d landmarks or flow based representations.
To increase the level of photorealism, a large number of works have extensively used discriminators as part of the losses \cite{ganimationPumarolaIJCV2019, wang2021one, headgan_Doukas_2021_ICCV, bounareli2022finding, yin2022styleheat, bounareli2023hyperreenact}, and some recent methods proposed the use of diffusion models \cite{facediffuser, thpad, stypulkowski2023diffused}. However, it is hard to ensure proper disentanglement between body, head motions, blinking, gaze and facial expressions when operating in the latent space of GANs \cite{goodfellow2014, karras2020analyzing} or generic diffusion models. Our method does not need to employ custom perceptual, gaze, identity preserving or lip syncing losses. %
Body motion and gestures have not been considered because of the lack of data and the difficulty of generating coherent video. We curate a large-scale dataset and propose a complete pipeline towards this problem. VLOGGER can generate coherent face and upper-body motion with a variety of expressions, head and body motion, gaze, eye blinking and accurate lip movement. %
Moreover, we show that our method is more expressive and robust across different diversity axis. %

\paragraph{Face Reenactment.} Video-based talking face generation aims to transfer the motion of a source video to a target person, and has been widely explored in the past \cite{zhou2019talking, zhang2019one,nirkin2019fsgan,thies2016face2face,hsu2022dual,yang2022face2face,bounareli2023hyperreenact, zhao2022thin, hong2022depth, wiles2018x2face,ha2020marionette}. Most methods rely on an intermediate representation, such as sparse or dense landmarks, semantic masks, 3d dense representations or warped features. In the 3d domain, several works have taken advantage of NeRF \cite{mildenhall2020nerf, barron2022mipnerf360} based solutions \cite{guo2021adnerf, liu2022semantic, yao2022dfa, genefaceICLR2023}. However, this requires a significant amount of frames of a target person talking, for retraining and  animating them.
This task is closely related to ours, and some previous works adapt these intermediate representations when considering audio as input. In our case, however, we aim to move forward from face-only videos and consider more diverse input samples, \eg containing body and hair motion.

\paragraph{Video Generation.} Also related to our work is the topic of video generation. This is a task that has been widely explored in the community, thus we only focus on the most related directions. With the success of text-to-image diffusion models \cite{dhariwal2021diffusion}, many works have also explored their extension to the video domain~\cite{kondratyuk2023videopoet,blattmann2023stable,bar2024lumiere,sora,ho2022imagen, lvdm, text2video-zero, makeyourvideo, phenaki} but most are limited in number of seconds or resolution.
Moreover, most previous works do not explicitly tackle humans despite the amount of data available. %
In our case, we extend current state-of-the-art image diffusion models to the temporal domain by adding spatio-temporal controls and propose an iterative outpainting procedure to generate videos of variable length.
While concurrent works explore similar network architectures~\cite{bar2024lumiere, sora} for more generic scenarios, our goal is to animate talking humans by parameterizing each frame with 1) dense renders of a posed 3D body model and 2) warped reference images. These controls make the generative process more stable as ablated in the experimental section.

\begin{figure*}[t]
\hspace{-0.8cm}
\includegraphics[width=1.15\linewidth, trim={0.2cm 0.3cm 0.3cm 0.2cm}, clip=true]{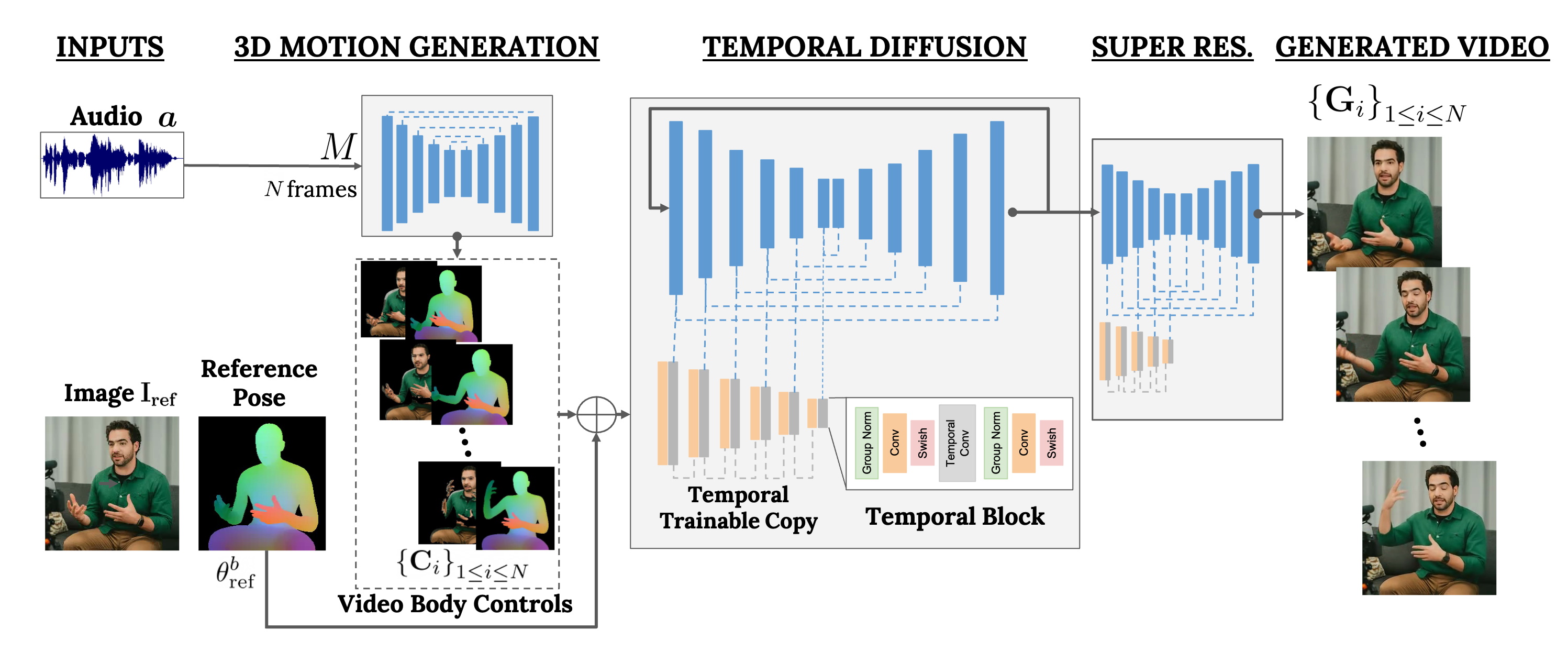} %
\caption{\textbf{High-level overview.} 
VLOGGER conditions the video generation process using a statistical 3D body model. 
Given an input image $\mathbf{I}_{\mathbf{ref}}$ (left), the predicted shape parameters encode the geometric properties of the target identity.
First, a network $M$ takes the Mel-Spectrogram $\mathbf{a}$ of an input speech and generates a sequence of 3D facial expressions $\left\{ \expression_{i} \right\}_{1 \leq i \leq N}$ and body poses $\left\{ \bodypose_{i} \right\}_{1 \leq i \leq N}$ for $N$ frames. We render dense representations of the moving 3D body to act as 2D controls $\left\{ \mathbf{C}_{i} \right\}_{1 \leq i \leq N}$ in the video generation stage (examples of controls in Sup. Mat.).
Together with the reference image of the subject, these are given as input to a temporal diffusion model and a super-resolution module, which are trained to generate a sequence of photorealistic reenactments $\left\{ \mathbf{G}_{i} \right\}_{1 \leq i \leq N} $ of the target identity. 
Implementation details in Sup. Mat.
}
\label{fig:diagram}
\end{figure*}

\section{Method}
\label{sec:method}

Our goal is to generate a photorealistic video $\video$ of variable length synthesizing a target human talking, with realistic head motion and gestures.  Our framework, which we call \Model, is illustrated in \cref{fig:diagram}. \Model is a two-stage pipeline based on stochastic diffusion models to represent the one-to-many mapping from speech to video.
The first network takes as input an audio waveform $\audio\in\mathbb{R}^{\videolength\samplerate}$ at sample rate $\samplerate$ to generate intermediate body motion controls $\controls$, which are responsible for gaze, facial expressions and 3D pose over the target video length $N$. The second network is a temporal image-to-image translation model that extends large image diffusion models, taking the predicted body controls to generate the corresponding frames. To condition the process to a particular identity, the network also takes a reference image of a person. %
We train \Model on our newly introduced \Dataset dataset (\S\ref{sec:dataset}). We describe both networks next.

\subsection{Audio-Driven Motion Generation}\label{sec:sec3.1}

\paragraph{Architecture.}
The first network of our pipeline $M$ is designed to predict the driving motion based on an input speech. We also consider input text through a text-to-speech model to convert inputs to waveforms~\cite{ttscloud}, and represent the resulting audio as standard Mel-Spectrograms. $M$ is based on a transformer architecture~\cite{vaswani2017attention} with four multi-head attention layers on the temporal dimension. We include positional encoding on the number of frames and diffusion step, and an embedding MLP for the input audio and the diffusion step.
At each frame, we use a causal mask to make the model attend only to previous frames. The model is trained using variable length videos to enable generation of very long sequences, as \eg in the TalkingHead-1KH Dataset \cite{talkinghead} (see \S\ref{sec:experiments}).

We rely on the estimated parameters of a statistical and expressive 3D body model~\cite{smplx, totalcapture, ghum2020, xavatar}
to produce intermediate control representations for the synthesized video. 
These models consider both facial expressions and body motion, opening the door for human synthesis with more expressive and dynamic gestures. 
We task the motion generation network to predict face and body parameters $M(\audio_i) = \{\expression_i, \Delta\bodypose_i\}$ based on the input audio $\audio_i$ in frame $i$. In particular, the model generates expression $\expression_i$ and residuals over body pose $\bodypose_i$.
By predicting displacements, \ie  $\Delta\bodypose_i$, we enable the model to take an input image with reference pose $\bodypose_{\text{ref}}$ for the target subject, and animate the person relatively with $\bodypose_i =\bodypose_{\text{ref}} + \Delta\bodypose_i$, for frames $1 \leq i \leq \videolength$.
The identity of the person in the geometric domain is modelled by the body shape code. 
During both training and testing, we use the estimated 3D shape parameters obtained by fitting the parametric body model to the input image. In order to leverage the 2D/3D predictions with CNN-based architectures, we pose the model using the predicted expression and pose parameters and rasterize the template vertex positions of the posed body as dense representations 
to obtain dense masks $\left\{ \mathbf{C}^{d}_{i} \right\}_{1 \leq i \leq \videolength} \in \mathbb{R}^{\height \times \width \times 3}$.
We also rasterize the semantic regions of the body, $\left\{ \mathbf{C}^{m}_{i} \right\}_{1 \leq i \leq \videolength} \in \{0, 1\}^{\height \times \width \times N_c}$, for $N_c$ different semantic classes. 

Furthermore, previous face reenactment works often rely on warped images~\cite{PECHead, sadtalker, audio2head, zhao2022thin}, yet these have been overlooked in diffusion-based architectures for human animation~\cite{hu2023animate, wang2023disco,chang2023magicdance}. We propose bridging the gap between these two representations and use warped images to guide the generative process, which we notice facilitates the task of the network and helps preserve subject identity (See \cref{tab:ablation_reenactment}).
We assign a pixel color to each body vertex that is visible in the reference image, and render the body in each new frame, obtaining partial warps $\left\{ \mathbf{C}^{w}_{i} \right\}_{1 \leq i \leq \videolength} \in \mathbb{R}^{\height \times \width \times 3}$. 
For all renders, the rasterization process assumes a full-perspective camera, with a diagonal field-of-view inferred from either the training video, or the reference image. For illustrations, please see \cref{fig:diagram}. We describe the temporal image diffusion model in the next section and in Sup. Mat. We also ablate the use of dense representations and warped images in the experimental section.

\paragraph{Loss functions.}
This model follows a diffusion framework which progressively adds Gaussian noise $\epsilon \sim \mathcal{N}(0, 1)$ to ground-truth samples $x_0 = \{\left\{ \expression_i, \Delta\bodypose_i \right\}\}_{1 \leq i \leq \videolength}$, with a conditional audio input $\audio$. The goal is to model the motion distribution of real heads and bodies, $x_0 \sim q(x_0|\audio)$, by training a denoising network $\epsilon_\phi$ that predicts the added noise from the noisy input $x_t$, where $t$ is an arbitrary diffusion step. In our case, we obtained better performance by directly predicting the ground-truth distribution
\begin{equation}
\loss_{\text{diff}} = \mathbb{E}_{x_0, t, \audio, \epsilon \sim \mathcal{N}(0, 1) }\Big[ \left\|x_0 - \epsilon_\phi(x_t, t, \audio) \right\|_{2}^{2}\Big].
\label{eq:loss_diffusion_motion}
\end{equation}
We also include an additional temporal loss to penalize prediction difference at consecutive frames, $\loss_{\text{temp}} = \left\| \epsilon_\phi(x_t, t, \audio)_{i+1} - \epsilon_\phi(x_t, t, \audio)_{i} \right\|_{2}^{2}$, for any given frame $i \in N$, and train the full model using a linear combination of both losses, \ie $\loss_{\text{diff}} + \lambda_{\text{temp}}\loss_{\text{temp}}$.
In practice, we use different temporal loss weights for expressions and body pose to ensure smoother motion for the head and hands while allowing larger dynamism for facial expressions.

\subsection{Generating Photorealistic Talking and Moving Humans }

\paragraph{Architecture.} 
Our next goal is to animate an input image $\bI_{\mathbf{ref}}$  of a person, such that it follows the previously predicted body and face motion, which is represented with semantic, sparse and dense masks $\controls$. 
Based on these image-based controls, we propose a temporally-aware extension of state-of-the-art diffusion models~\cite{imagen}. %
Inspired by ControlNet~\cite{zhang2023adding}, we freeze the initial trained model and make a zero-initialized trainable copy of its encoding layers, which take the input temporal controls $\controls$. We interleave 1d convolutional layers in the temporal domain, after the first layer of each downsampling block and before the second GroupNorm activation, as shown in \cref{fig:diagram}.
The network is trained by taking $N$ consecutive frames and controls, and tasked to generate short clips of the reference person animated according to the input controls. 

\paragraph{Training.} 
We train our method on the \Dataset dataset, which consists of full-length videos of unique human subjects. Because, during training, the network takes a sequence of consecutive frames and an arbitrary reference image $\bI_{\mathbf{ref}}$ of the person, we theoretically can assign any video frame as reference. In practice, we sample the reference to be farther away (temporally) from the target clip, as closer examples trivialize the training and provide less generalization potential. 
The network is trained in two stages by first learning the new control layers~\cite{zhang2023adding} on single frames, and later training on videos by adding the temporal components. This enables using a large batch size in the first stage and learning the head reenactment task faster. 
We train the image models with learning rate 5e-5, for $400k$ steps with batch size 128 in both stages. 
We ablate the effect of this training schedule in Table~\ref{tab:ablation_reenactment} and more details about the training procedure are provided in \suppmat.

\paragraph{Loss functions.}
Similar to the previous section and the loss described in \cref{eq:loss_diffusion_motion}, we follow a diffusion process in which we add noise $\epsilon^{I}$ to the ground-truth images $\bI$. 
We base our work on a version of Imagen~\cite{imagen} trained on internal data sources,
which predicts the added noise $\epsilon^I$
\begin{equation}
\loss^{I}_{\text{diff}} = \mathbb{E}_{x^{I}_0, t, \controls, \epsilon^{I} \sim \mathcal{N}(0, 1) }\Big[ \left\| \epsilon^{I} - \epsilon^{I}_{\phi}(x^{I}_t, t, \mathbf{C}) \right\|_{2}^{2}\Big].
\end{equation}
\noindent\paragraph{Super Resolution.} While the previous approach is resolution independent, we generate base videos at $128\times128$ resolution, and use a cascaded diffusion approach to extend the temporal conditioning in two super-resolution variants for higher quality video at  $256\times256$ or $512\times512$. The  generated images are denoted as $\left\{\mathbf{G}_{i}\right\}_{1\leq i \leq \videolength}$. High resolution examples are shown in \cref{fig:teaser} and \cref{fig:results_qualitative}. %

\paragraph{Temporal outpainting during inference.} The proposed temporal diffusion model is trained to generate only a fixed number of frames $N$, so it is not obvious how to extend it to variable length videos. Most previous diffusion-based video generation methods are limited to short clips~\cite{ho2022video,text2video-zero,makeyourvideo} or rely on smoothly generated intermediate token representations~\cite{phenaki}, but without guarantees of smooth changes in the pixel domain. Here, we explore the idea of temporal outpainting: we first generate $N$ frames, and then we iteratively outpaint $N' < N$ frames based on the previous $N - N'$. The amount of overlap between two consecutive clips, \ie $N - N'$ is chosen as a trade-off between quality and running time. We use DDPM to generate each video clip, and show that such approach can scale to thousands of frames. For details, see the ablation in \cref{tab:ablation_video}, where we validate the main design choices and show that our final network can generate realistic and temporally coherent videos of humans. 

\subsection{\Dataset Dataset}
\label{sec:dataset}

We curate the \Dataset Dataset from a large repository of internal videos that contain a single speaker, mostly facing the camera, from the torso up, communicating mostly in English.
The videos contain $240$ frames at $24$ fps ($10$ seconds clips), with audio at $16$ kHz.

With the goal of modelling full-body communicating humans, we estimate 3d body joints and hands and fit a statistical articulated 3D body model by minimizing the projection error and temporal difference between consecutive frames. 
We filter out videos where the background changes meaningfully, the face or body have been only partially detected or their estimations are \textit{jittery}, where hands are completely undetected (\eg in cases of humans grasping and manipulating objects), or the audio is of low quality.
This process resulted in a training set of more than 8M seconds (2.2K hours) and 800K identities, and a test set of 120 hours and ${\sim}4$K identities, making it the largest dataset used to date in terms of identities and length, at higher resolution. Moreover, the \Dataset dataset contains a wide diversity of subjects (\eg skin tone, age), viewpoints or body visibility. Statistics and a broader comparison to existing datasets are provided in \suppmat.
We aim to release the curated video ids, face fits and estimated body pose to the broader research community.

\section{Experiments}
\label{sec:experiments}

\paragraph{\bf Data and Training. }
We train \Model on the \Dataset dataset as described in \cref{sec:dataset}, at a base resolution of $128\times128$ and cascade resolutions at $256\times256$ and $512\times512$. Evaluation is performed on the test sets of the HDTF~\cite{hdtf}, TalkingHead-1KH \cite{talkinghead} and \Dataset.
We also ablate the performance of our method in different scenarios on the \Dataset dataset and report its performance against baselines across several diversity metrics, such as age, perceived gender, or skin tone.

\paragraph{\bf Baselines.} 
We compare against several state-of-the-art methods, i.e. ~\cite{makeittalk, audio2head, wang2022one, sadtalker, styletalk}.
Note that, unlike our method, all baselines require cropping the face region, as they can detect and animate only the head.

\paragraph{\bf Metrics.} 
We rely on a combination of metrics to evaluate image quality, lip sync, temporal consistency, and identity preservation of the generated videos. 
For image quality, the FID score~\cite{fid} measures the distance between ground-truth and generated image distributions, while the Cumulative Probability of Blur Detection (CPBD)~\cite{cpbd1,cpbd2} and Natural Image Quality Evaluator (NIQE)~\cite{niqe} validate the quality of generated images.
Following the literature in talking face generation, we next estimate face landmark coordinates and report the difference in mouth vertex position (LME) to measure lip sync quality. We also report the LSE-D~\cite{syncnet} score. Similarly, we report the jitter (or \textit{jerk}) error following~\cite{yi2021transpose} to measure the temporal smoothness in generated videos. We also provide the standard deviation of the expression parameters predicted from generated videos, to assess diversity in terms of expression and gaze, given that speech-to-video is not always a one-to-one mapping and it is important to generate a distribution of realistic videos. Regarding diversity of body and hand motion, VLOGGER is the first model to consider gestures, and we assess this qualitatively.

\subsection{Ablation Study}

We ablate our main design choices extensively in Tables~\ref{tab:ablation_video} and~\ref{tab:ablation_reenactment}. \cref{tab:ablation_video} summarizes the most representative metrics for the full method (last row) and each row represents the effect of changing one feature (\eg not using a temporal loss when training the motion predictor). \cref{tab:ablation_reenactment} validates the importance of the 2d controls used to generate videos. 
We discuss the results next.

\begin{table}[t]
\centering
\resizebox{.72\linewidth}{!}{%
\begin{tabular}{r c c c }
 & FID~\cite{fid} $\downarrow$ & LME [mm] $\downarrow$ & Jitter [$\text{mm}/\text{s}^{3}$] $\downarrow$ \\
\cmidrule(lr){2-4}
Metrics in the final video & \multicolumn{3}{c}{Motion Generation} \\
\cmidrule(lr){1-1} \cmidrule(lr){2-4}
\small{Not predicting $\Delta$ over body pose} & 52.27 & 4.22 & 6.56 \\
\small{Not training with temporal loss} & 16.56 & 3.18 & 4.64 \\
\small{Not using classifier-free guidance} & 16.54 & 3.32 & \textbf{3.49} \\
\cmidrule(lr){1-1} \cmidrule(lr){2-4}
& \multicolumn{3}{c}{Temporal Diffusion Model} \\
\cmidrule(lr){1-1} \cmidrule(lr){2-4}
\small{No body controls (Only renders of head area)} & 16.95 & 3.10 &  4.45 \\ %
\small{No temporal outpainting during inference} & 15.88 & 3.25 & 3.70 \\
\small{25\% outpainting overlap during inference} & 15.90 & 3.23 & 3.61 \\
\cmidrule(lr){1-1} \cmidrule(lr){2-4}
\small{Full model} & \textbf{15.36} & \textbf{3.06} & 3.58 \\
\cmidrule(lr){1-1} \cmidrule(lr){2-4}
\end{tabular}}
\caption{{\bf Ablation study of the main design choices in \Model} evaluated on the \Dataset~Dataset, where we report the most representative metrics to validate image quality through the FID~\cite{fid} score, expressiveness and lip sync quality via landmark error (LME), and temporal consistency based on face vertex jitter. The first part shows that the temporal loss and classifier-free guidance lead to the best performance in image quality and LME (full model in last row for comparison). 
The second part summarizes improvements for design choices in the temporal diffusion model. The final pipeline benefits from taking body controls, and the proposed temporal outpainting (50\% overlap in the full model) results in the best temporal consistency. We noticed the model plateaus with more overlap.
}
\label{tab:ablation_video}
\end{table}

\begin{table}[t]
\centering
\resizebox{1\linewidth}{!}{%
\begin{tabular}{r c c c c c c c c c c}
& \multicolumn{2}{c}{Face} & \multicolumn{2}{c}{Body} & \multicolumn{2}{c}{Hands} & \multicolumn{4}{c}{Full Image}
\\
\cmidrule(lr){2-3} \cmidrule(lr){4-5} \cmidrule(lr){6-7} \cmidrule(lr){8-11}
& PSNR $\uparrow$ & L1 $\downarrow$ & PSNR $\uparrow$ & L1 $\downarrow$ & PSNR $\uparrow$ & L1 $\downarrow$ & PSNR $\uparrow$ & SSIM $\uparrow$ & LPIPS $\downarrow$ & L1 $\downarrow$ \\
\cmidrule(lr){1-1} \cmidrule(lr){2-3} \cmidrule(lr){4-5} \cmidrule(lr){6-7} \cmidrule(lr){8-11}
\small{Using 2D body keypoints} & 20.5 & .0591 & 17.9 &.0778  & 17.8 & .0763 & 19.8 & .702 & 0.138 & .0564 \\
\small{Using Dense Body Representation} & 20.4 & .0604 & 18.3 & .0750 & 18.2 & .0744 & 20.1 & .719 & 0.128 & .0548 \\
\small{+ Warped Image Based on Body Model} & 21.6 & .0517 & 19.3 & .0668 & 19.1 & .0680 & 20.7 & .722 & 0.113 & .0496 \\
\small{+ Training Schedule (Full model)} & \textbf{22.2} & \textbf{.0468} & \textbf{20.2} & \textbf{.0594} & \textbf{20.0} & \textbf{.058} & \textbf{21.6} & \textbf{.76} & \textbf{.095} & \textbf{.0447 }\\
\cmidrule(lr){1-1} \cmidrule(lr){2-3} \cmidrule(lr){4-5} \cmidrule(lr){6-7} \cmidrule(lr){8-11}
\end{tabular}}
\caption{
{\bf Ablation of 2d controls in video generation,} in the \Dataset Dataset. 
We ablate different 2d controls considered in concurrent works, such as driving 2d skeleton~\cite{wang2023disco, hu2023animate}, dense body representations~\cite{xu2023magicanimate} or our proposed controls which include dense representations and warped images. 
In this experiment, we take the first image and animate the rest of the video following the original motion, reporting average image similarity metrics average and per body part. All variants are trained on the same data.
}
\label{tab:ablation_reenactment}
\end{table}

\begin{table*}[t]
\centering
\resizebox{\linewidth}{!}{%
\begin{tabular}{r c c c c c c c c c}
& \multicolumn{9}{c}{HDTF Dataset~\cite{hdtf}} \\
\cmidrule(lr){3-10}
& \multicolumn{3}{c}{Photorealism} & \multicolumn{2}{c}{Lip Sync} & \multicolumn{1}{c}{Diversity} & \multicolumn{2}{c}{Identity Preserv.} & Temp. Consist.\\
\cmidrule(lr){2-4} \cmidrule(lr){5-6} \cmidrule(lr){7-7} \cmidrule(lr){8-9} \cmidrule(lr){10-10}
& FID~\cite{fid} $\downarrow$ & CPBD~\cite{cpbd2} $\uparrow$ & NIQE~\cite{niqe} $\downarrow$ & LSE-D~\cite{syncnet} $\downarrow$ & LME [mm] $\downarrow$ & Expression $\uparrow$ & Head Err. $\downarrow$ & ArcFace~\cite{deng2019arcface} $\downarrow$ & Jitter [$\text{mm}/\text{s}^{3}$] $\downarrow$ \\
\cmidrule(lr){2-4} \cmidrule(lr){5-6} \cmidrule(lr){7-7} \cmidrule(lr){8-9} \cmidrule(lr){10-10}
\small{Groundtruth}  & 0.00 & 0.562 & 6.31 & 7.79 & 0.0 & 0.401 & 0.00 & 0.00 & 5.19 \\
\small{MakeItTalk \cite{makeittalk}}  & 22.63 & 0.428 & 6.65 & 8.30 & 3.26 & 0.364 & 0.911 & 0.828 & 6.21 \\
\small{Audio2Head \cite{audio2head}}  & 19.58 & 0.512 & \underline{6.41} & \textbf{7.55} & 3.08 & \underline{0.415} & 0.896 & 1.92 & 6.15 \\
\small{Wang~\etal~\cite{wang2022one}}  & 21.23 & 0.428 & 7.71 & 8.04 & 4.48 & 0.365 & 1.37 & 2.52 & 6.46 \\
\small{SadTalker \cite{sadtalker}} & \underline{19.44} & \underline{0.520} & 6.48 & \underline{7.73} & \textbf{3.01} & 0.287 &  
 \underline{0.880} & 0.874 & 5.51 \\
\small{StyleTalk \cite{styletalk}}  & 34.16 & 0.472 & 6.47 & 7.87 & 3.79 & \textbf{0.416} & 1.14 & \textbf{0.692} & \textbf{4.34} \\
\small{Ours} &
 \textbf{18.98} & 
 \textbf{0.621} &  
 \textbf{5.92} & 8.10 &  
 \underline{3.05} & 0.397 & \textbf{0.877} &  \underline{0.759} &  \underline{5.05} \\ 
\cmidrule(lr){2-10}
\small{Ours (Best of 3)}  & - & 0.628 & 5.64 & 7.43 & 2.95 & 0.425 & 0.829 &  0.706 & 4.75 \\
\small{Ours (Best of 5)}  & - & 0.631 & 5.53 & 7.22 & 2.91  &  0.436 & 0.814 & 0.687 & 4.67 \\
\small{Ours (Best of 8)}  & - & \textbf{0.634} & \textbf{5.44} & \textbf{7.04} & \textbf{2.84} & \textbf{0.448} & \textbf{0.800}  & \textbf{0.677} & \underline{4.58}  \\
\cmidrule(lr){2-10}\morecmidrules\cmidrule(lr){2-10}
& \multicolumn{9}{c}{TalkingHead-1KH Dataset~\cite{talkinghead}} \\
\cmidrule(lr){3-10}
& \multicolumn{3}{c}{Photorealism} & \multicolumn{2}{c}{Lip Sync} & \multicolumn{1}{c}{Diversity} & \multicolumn{2}{c}{Identity Preserv.} & Temp. Consist.\\
\cmidrule(lr){2-4} \cmidrule(lr){5-6} \cmidrule(lr){7-7} \cmidrule(lr){8-9} \cmidrule(lr){10-10}
&  FID~\cite{fid} $\downarrow$ & CPBD~\cite{cpbd2} $\uparrow$ & NIQE~\cite{niqe} $\downarrow$ & LSE-D \cite{syncnet} $\downarrow$ & LME [mm] $\downarrow$ & Expression $\uparrow$ & Head Err. $\downarrow$ & ArcFace \cite{deng2019arcface}$\downarrow$ & Jitter [$\text{mm}/\text{s}^{3}$] $\downarrow$ \\
\cmidrule(lr){2-4} \cmidrule(lr){5-6} \cmidrule(lr){7-7} \cmidrule(lr){8-9} \cmidrule(lr){10-10}
\small{Groundtruth} & 0.00 & 0.512 & 7.27 & 8.70 & 0.0 & 0.452 & 0.00 & 0.00 & 3.91 \\
\small{MakeItTalk \cite{makeittalk}}  & 34.84 & \underline{0.493} & 7.86 & 10.48 & 3.50 & 0.382 & \underline{1.20} & 0.909 & 4.69 \\
\small{Audio2Head \cite{audio2head}}  & 46.49 & 0.475 & 7.55 & \underline{9.38} & 4.33 & \textbf{0.494} & 1.47 & 2.01 & 4.66 \\
\small{Wang~\etal~\cite{wang2022one}}  & 34.52 & 0.440 & 8.61 & 10.18 & 3.49 & 0.338 & 1.48 & 2.93 & 4.70 \\
\small{SadTalker \cite{sadtalker}}  & \underline{31.45} & 0.482 & \underline{7.46} & \textbf{8.17} & \textbf{3.10}  & 0.347 & 1.21 & 0.961 & 4.26\\
\small{StyleTalk \cite{styletalk}} & 38.98 & 0.468 & 7.96 & 9.46  & 3.44  & 0.421 & 1.29 & \textbf{0.663} & \textbf{3.19} \\
\small{Ours}  & \textbf{28.94} & \textbf{0.575} & \textbf{6.91} & 9.40 & \underline{3.33} & \underline{0.436} & \textbf{1.05} &  \underline{0.881} & \underline{4.16} \\ 
\cmidrule(lr){2-10}
\small{Ours (Best of 3)}  & - & 0.582 & 6.33 & 8.969 & 3.07 & 0.448 & 1.03 & 0.853 & 3.68 \\
\small{Ours (Best of 5)}  & - & 0.585 & 6.21 & 8.93 & 2.96 & 0.455 & 1.01 & 0.833 & 3.57 \\
\small{Ours (Best of 8)}  & - & \textbf{0.589} & \textbf{6.08} & \underline{8.90}  & \textbf{2.94} & \underline{0.469} & \textbf{0.99} & \underline{0.813} &  \underline{3.56} \\
\cmidrule(lr){2-10}
\end{tabular}}
\caption{\textbf{Quantitative evaluation on the HDTF and TalkingHead-1KH Datasets.} We measure the capacity of our model to generate realistic talking heads in multiple metrics. 
\Model achieves the highest visual quality with highest identity preservation summarized in several metrics, while obtaining expression diversity and temporal consistency close to the groundtruth videos.
Regarding lip sync quality, all methods obtain comparable scores. 
To demonstrate the diversity generated by \Model, we also report the improvement in performance when generating 3, 5 or 8 videos (Except for FID which measures a similarity within an image distribution). 
Results are consistent for all metrics on both datasets. 
}
\label{tab:quant_full}
\end{table*}

\paragraph{\bf Motion generation.} In the upper-part of \cref{tab:ablation_video} we show the drop in temporal consistency when not using temporal loss or not predicting $\Delta$ (See Sec~\ref{sec:sec3.1}). The network gains in smoothness and stability when predicting a residual over body motion, resulting in overall higher image quality. We also show the positive use of classifier-free guidance (discussed in Sup. Mat.) regarding LME and FID~\cite{fid}.

\paragraph{\bf Video Generation.} The lower-part of \cref{tab:ablation_video} ablates the design choices on the temporal video generation model.
First, it validates the effectiveness of the proposed outpainting procedure, which not only supports variable-length video generation, but also ensures smoothness and low jitter. Our final model has an overlap of 50\% between generated and given frames, and plateaus at larger values, but obtains a noticeable improvement with respect to a smaller overlap (25\%), or no outpainting. The model also performs better with body pose control.

\paragraph{\bf Effect of 2d controls in video generation.} 
We finally ablate the importance of the different representations used to guide the video generation process in \cref{tab:ablation_reenactment}, by reenacting test set samples with their groundtruth motion and reporting image reconstruction metrics. We explore 2d landmarks, dense representations and our final proposed controls, which combine dense body representations and reference partial views warped from the reference input image. The latter eases the task of the network significantly and leads to the best results. Moreover, we obtain an additional boost in performance with the training schedule described in Section~\ref{sec:method} (and in Sup. Mat.), of first training in single images and later finetuning the temporal layers in videos.

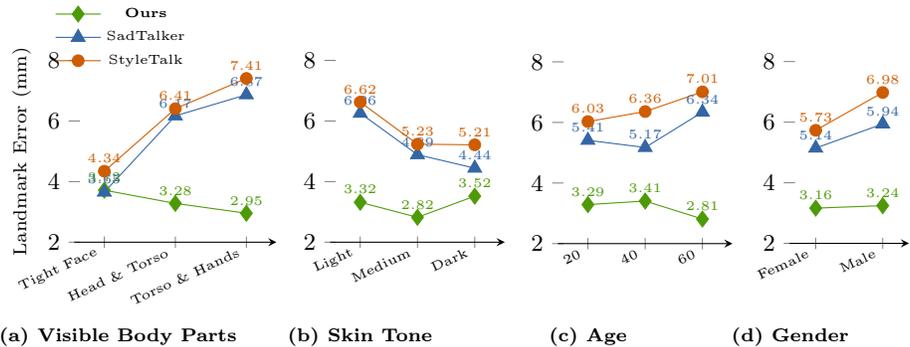
\begin{figure*}[t]
  \captionsetup[subfigure]{labelformat=empty}
  \footnotesize
  \hspace{-0.65cm}
  \subfloat[\textbf{(a) Visible Body Parts} \label{fig:diversity_attributes}] {
    \begin{tikzpicture}
        \begin{axis}[
            height=4cm, width=0.35\textwidth,
            ymin=2.0, ymax=8.,
            ylabel={\scriptsize{Landmark Error (mm)}},
            y axis line style={opacity=0},
            axis lines=left,
            symbolic x coords={Tight Face, Head \& Torso, Torso \& Hands},
            xtick=data,
            x tick label style={font=\tiny, anchor=east, rotate=25, align=right, text width=1.6cm},
            enlarge x limits={abs=0.4cm},
            nodes near coords,
            every node near coord/.append style={font=\tiny},
            legend style={
                at={(0.24, 1.351)},
                anchor=north,
                draw=none,
                fill=none,
                font=\tiny,
                },
            ]
            \addplot+[chameleon3, mark=diamond*,mark options={scale=1.5, fill=chameleon3}] coordinates{ (Tight Face, 3.7161216605454683) (Head \& Torso, 3.280926262959838) (Torso \& Hands, 2.9545046854764223)};
            \addplot+[skyblue2, mark=triangle*,mark options={scale=1.5, fill=skyblue2}]
            coordinates{ (Tight Face, 3.6520701833069324) (Head \& Torso, 6.169890519231558) (Torso \& Hands, 6.865944247692823)};
            \addplot+[orange3, mark=*,mark options={scale=1.1, fill=orange3}]
            coordinates{ (Tight Face, 4.342430736869574) (Head \& Torso, 6.406489294022322) (Torso \& Hands, 7.405478041619062)};
            \legend{\bf{Ours}, SadTalker, StyleTalk},
        \end{axis}
    \end{tikzpicture}
  }\hspace{-0.95cm}
  \subfloat[\textbf{(b) Skin Tone} \label{fig:diversity_skin}] {
    \begin{tikzpicture}
        \begin{axis}[
            height=4cm, width=0.32\textwidth,
            ymin=2.0, ymax=8.,
            y axis line style={opacity=0},
            axis lines=left,
            symbolic x coords={
               Light, Medium, Dark},
           xtick=data,
           x tick label style={font=\tiny, anchor=east, align=right, rotate=25, text width=1.6cm},
           enlarge x limits={abs=0.4cm},
           nodes near coords,
           every node near coord/.append style={font=\tiny},
            ]
            \addplot+[chameleon3, mark=diamond*,mark options={scale=1.5, fill=chameleon3}] coordinates{ (Light, 3.3169679809361696) (Medium, 2.82063870690763) (Dark, 3.519603982567787) };
            \addplot+[skyblue2, mark=triangle*,mark options={scale=1.5, fill=skyblue2}]
            coordinates{ (Light, 6.257820874452591) (Medium, 4.888787399977446) (Dark, 4.443218931555748) };
            \addplot+[orange3, mark=*,mark options={scale=1.1, fill=orange3}]
            coordinates{ (Light, 6.623473018407822) (Medium, 5.2348230965435505) (Dark, 5.213775672018528) };
        \end{axis}
    \end{tikzpicture}
  }\hspace{-0.95cm}   %
  \subfloat[\textbf{(c) Age} \label{fig:diversity_age}] {
    \begin{tikzpicture}
        \begin{axis}[
            height=4cm, width=0.32\textwidth,
            ymin=2.0, ymax=8.,
            y axis line style={opacity=0},
            axis lines=left,
            symbolic x coords={
               20, 40, 60},
           xtick=data,
           x tick label style={font=\tiny, anchor=east, rotate=25, align=right, text width=1.6cm},
           enlarge x limits={abs=0.4cm},
           nodes near coords,
           every node near coord/.append style={font=\tiny},
            ]
            \addplot+[chameleon3, mark=diamond*,mark options={scale=1.5, fill=chameleon3}] coordinates{ (20, 3.2872885931283236) (40, 3.405739087611437) (60, 2.81116203404963) };
            \addplot+[skyblue2, mark=triangle*,mark options={scale=1.5, fill=skyblue2}]
            coordinates{ (20, 5.410614423453808) (40, 5.173202138394117) (60, 6.339764688163996) };
            \addplot+[orange3, mark=*,mark options={scale=1.1, fill=orange3}]
            coordinates{ (20, 6.028007250279188) (40, 6.358931306749582) (60, 7.009648252278566) };
        \end{axis}
    \end{tikzpicture}
  }\hspace{-0.95cm}
  \subfloat[\textbf{(d) Gender} \label{fig:diversity_gender}] {
    \begin{tikzpicture}
        \begin{axis}[
            height=4cm, width=0.26\textwidth,
            ymin=2.0, ymax=8.,
            y axis line style={opacity=0},
            axis lines=left,
            symbolic x coords={
               Female, Male},
           xtick=data,
           x tick label style={font=\tiny, anchor=east, rotate=25, align=right, text width=1.6cm},
           enlarge x limits={abs=0.35cm},
           nodes near coords,
           every node near coord/.append style={font=\tiny},
           legend pos=north east,
        ]
            \addplot+[chameleon3, mark=diamond*,mark options={scale=1.5, fill=chameleon3}] coordinates{ (Female, 3.1618946231901646) (Male, 3.2449180725961924)};
            \addplot+[skyblue2, mark=triangle*,mark options={scale=1.5, fill=skyblue2}]
            coordinates{ (Female, 5.144426133483648) (Male, 5.940518341958523)};
            \addplot+[orange3, mark=*,mark options={scale=1.1, fill=orange3}]
            coordinates{ (Female, 5.731677170842886) (Male, 6.979325320571661)};
        \end{axis}
    \end{tikzpicture}
  }
  \caption{Our model and closest competitors across {\bf different perceived attributes}, such as skin tone, gender and age, on the test set of the \Dataset dataset. Our model leverages priors from large pre-trained diffusion models and our proposed large-scale dataset. Thus, in contrast to other methods, it manages to perform consistently across all categories, showing little to no bias. We also show in \subref{fig:diversity_attributes} that our model is capable of animating humans in images at a wide range of viewpoints, instead of cropping tight bounding boxes around the face. 
  }
\label{fig:diversity}
\end{figure*}

\subsection{\bf Quantitative Results}

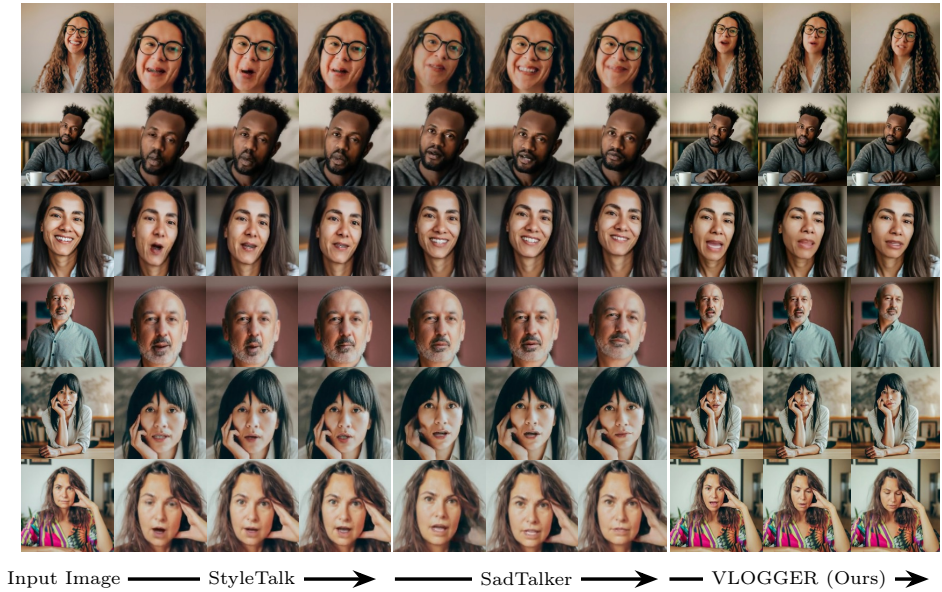
\begin{figure*}[t]
\begin{center}
\includegraphics[width=\linewidth]{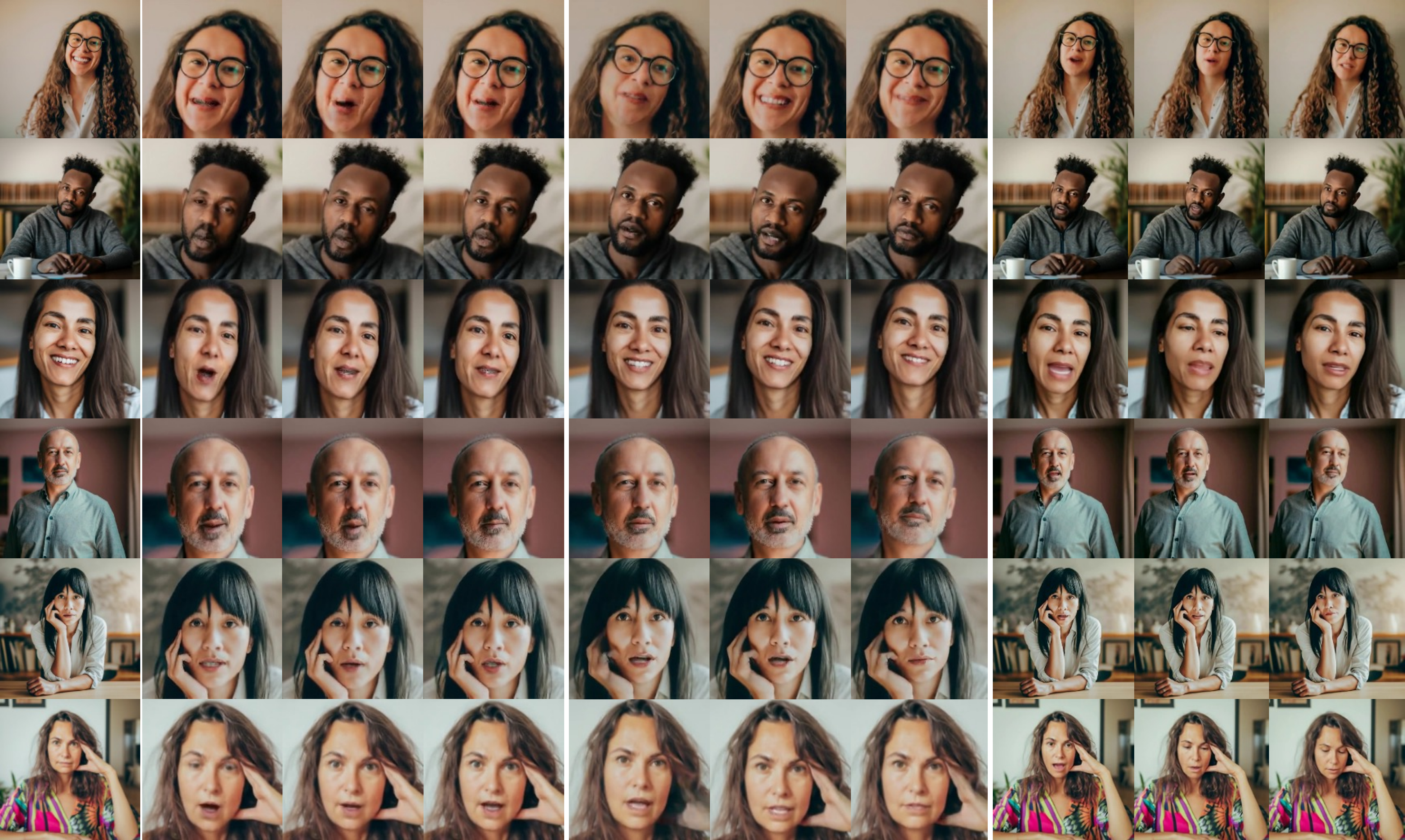}
\end{center}
\vspace{-0.25cm}
\begin{tikzpicture}
  \hspace{-0.3cm}
  \draw node (0,0) {\scriptsize{Input Image}};
  \draw[-{Stealth[black]},line width=1.3pt] (0.85,0)   -- (4.15,0) 
  node [midway,fill=white] {\scriptsize{StyleTalk}};
  \draw[-{Stealth[black]},line width=1.3pt] (4.4,0)   -- (7.9,0) 
  node [midway,fill=white] {\scriptsize{SadTalker}};
  \draw[-{Stealth[black]},line width=1.3pt] (8.05,0)   -- (11.5,0)
  node [midway,fill=white] {\scriptsize{\Model (Ours)}};
\end{tikzpicture}\\
\caption{\textbf{Qualitative comparison} showing input images (left) and generated frames. %
Baselines typically maintain the expression along the whole sequence, and require cropping the head~\cite{sadtalker, styletalk, wang2022one}.
 In contrast, VLOGGER generates changes in the visible areas when considering faces (third row) but also visible upper-body (fifth row).
This figure shows animated faces, but examples with gestures are shown in \cref{fig:teaser} and Sup. Mat.
\label{fig:results_qualitative}
}
\end{figure*}

\paragraph{\bf Talking Head Generation.} 
\cref{tab:quant_full} summarizes the performance of \Model against previous state-of-the-art methods on the task of audio-driven video generation. We report results on the HDTF Dataset~\cite{hdtf}, a large scale dataset, but with a low number of identities (300) subjects and somewhat limited viewpoint variability, and on the TalkingHead-1KH Dataset~\cite{talkinghead}. 
Talking head generation is a challenging task with several desirable properties, assessed by different metrics. Noticeably, there is a trade-off between image quality, diversity and identity preservation. \Model comes close to the amount of expression diversity present in real videos while achieving the highest image quality and identity preservation, with second lowest motion jitter after StyleTalk \cite{styletalk}, which introduces very little face motion (see \cref{fig:results_qualitative}).
The temporal consistency validates the contribution of our temporal layer and the outpainting procedure, while still leveraging the high-quality image generation capabilities of state-of-the-art diffusion models. 
All methods obtain comparable Lip Sync scores, and results are consistent for all metrics on both datasets evaluated. 
We also evaluate our method with different number of samples produced (3, 5 or 8) by selecting the best performing video per subject,
leading to significantly improved performance with growing number of samples. These support the generative properties of \Model, showing its capacity to generate different samples per subject.
Also, note that these consider images of faces only, while our goal is to model visible body parts including hands. While no baselines consider body or gestures, we ablate our design choices in this regard in Tables~\ref{tab:ablation_video} and \ref{tab:ablation_reenactment}.

In \cref{fig:diversity}, we showcase our fairness and generalization capabilities (in part due to the scale and diversity of our training set), by running comparisons to other methods across several perceived attributes. Previous works exhibit a clear performance degradation for different classes (\eg light vs dark skin, young vs old, \etc), and do not generalize to videos with visible torsos or hands. In contrast, \Model exhibits fairly low bias on all the evaluated axes. 
We hope that the release of \Dataset will enable the community to address critical fairness issues and further advance the state-of-the-art.

\begin{figure}[t!]
\centering
\includegraphics[width=\linewidth]{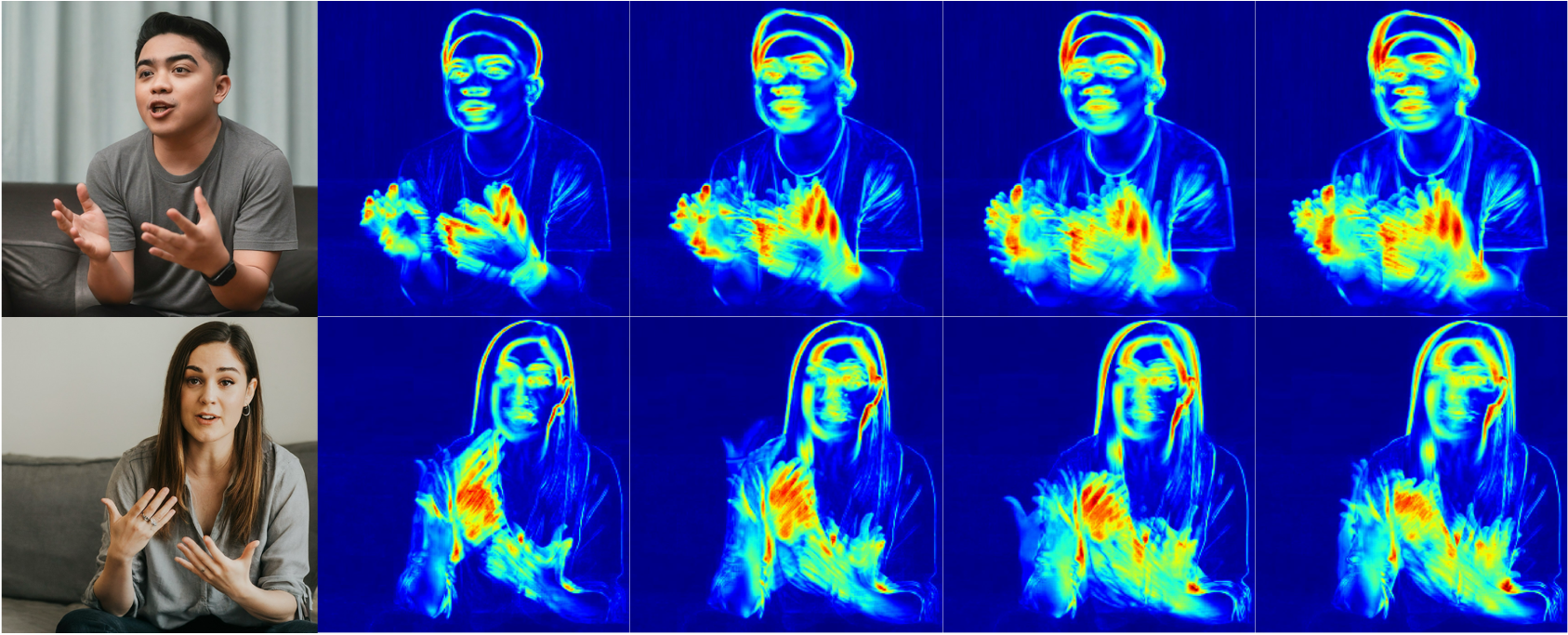}
\put(-333,-7){\scriptsize{Input image}}
\put(-56,-7){\scriptsize{Pixel Diversity}}
\\
\vspace{-0.28cm}
\begin{tikzpicture}
  \draw[-{Stealth[black]},line width=1.3pt] (0.0,0) -- (7.4,0) ;
\end{tikzpicture}
\caption{\textbf{Showcasing model diversity}. 
\Model~is stochastic and can generate a variety of videos for the same subject. Given the subject images and an input speech, columns 2-5 show the deviation in pixel color after 1-4 seconds respectively, obtained from 24 generated videos.
After only one second (second col.) the model already shows great diversity in hand pose and facial expressions, with all videos of good visual quality.
}
\label{fig:results_stochasticity}
\end{figure}
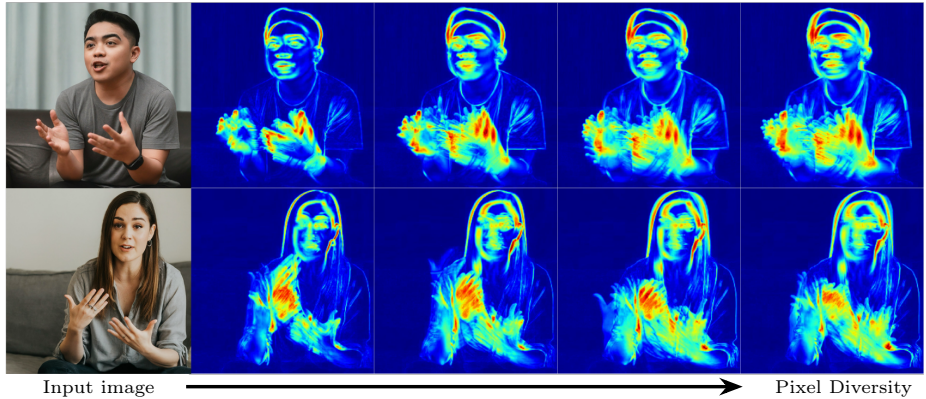

\subsection{\bf Qualitative Results} 

We show qualitative results in \cref{fig:results_qualitative} against the most recent and high-performing baselines on images in-the-wild.
Most previous works have limited generative capacity, which makes it difficult to generate parts occluded in the reference image (\eg if the teeth were obscuring the mouth interior, they will persist across the generated video). In contrast, our model is able to generate more diverse expressions and correctly inpaint occluded regions of moving heads.

\paragraph{\bf Sample diversity.} Since \Model is stochastic, we can generate multiple motions and videos given the same input audio/text, as illustrated in  \cref{fig:results_stochasticity}. From the first row, it can be seen that while the background is almost static, the face, hair, gaze and body motion feature an increasing amount of change as the video temporally unfolds.

\begin{figure}[t]
\centering
\includegraphics[width=1\linewidth]{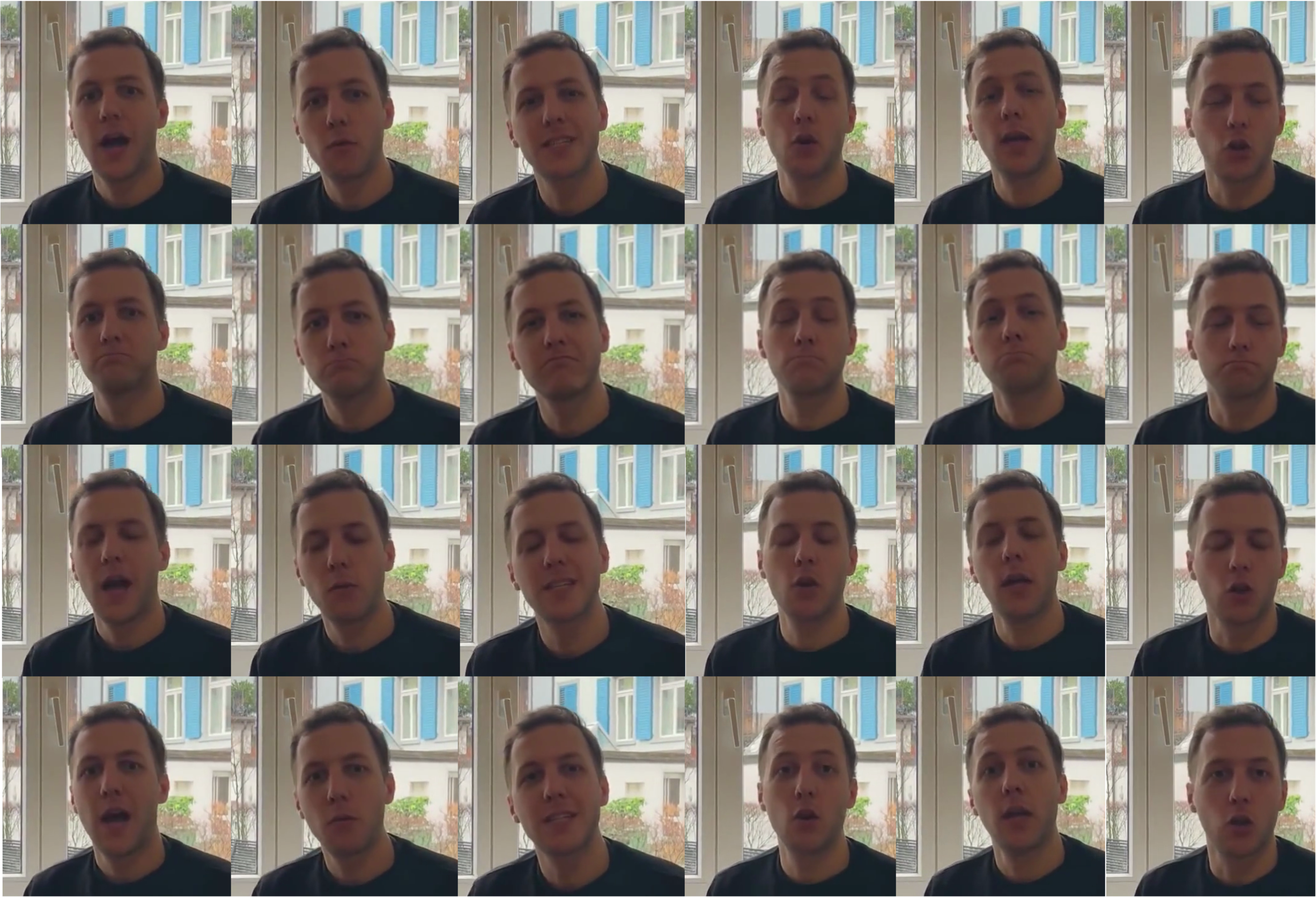}
\put(-356,184){\rotatebox{90}{{\scriptsize{Input Video}}}}
\put(-356,122){\rotatebox{90}{\scriptsize{Mouth Closed}}}
\put(-356,67){\rotatebox{90}{\scriptsize{Eyes Closed}}}
\put(-356,6){\rotatebox{90}{\scriptsize{Not Blinking}}}
\\
\begin{tikzpicture}
  \draw[-{Stealth[black]},line width=1pt] (0.0,0)   -- (11.0,0) ;
\end{tikzpicture}\\
\caption{
\textbf{Video editing results}. Given an input video (first row), we define new face expressions to change the mouth (second row), eyes (third row) or keep eyes open during the whole video (fourth row).
The temporal inpainting mask is defined from the changing parts of the body automatically. Best seen in Sup. Mat.
}
\label{fig:results_editing}
\end{figure}

\begin{figure}[t!]
\centering
\includegraphics[width=.65\linewidth]{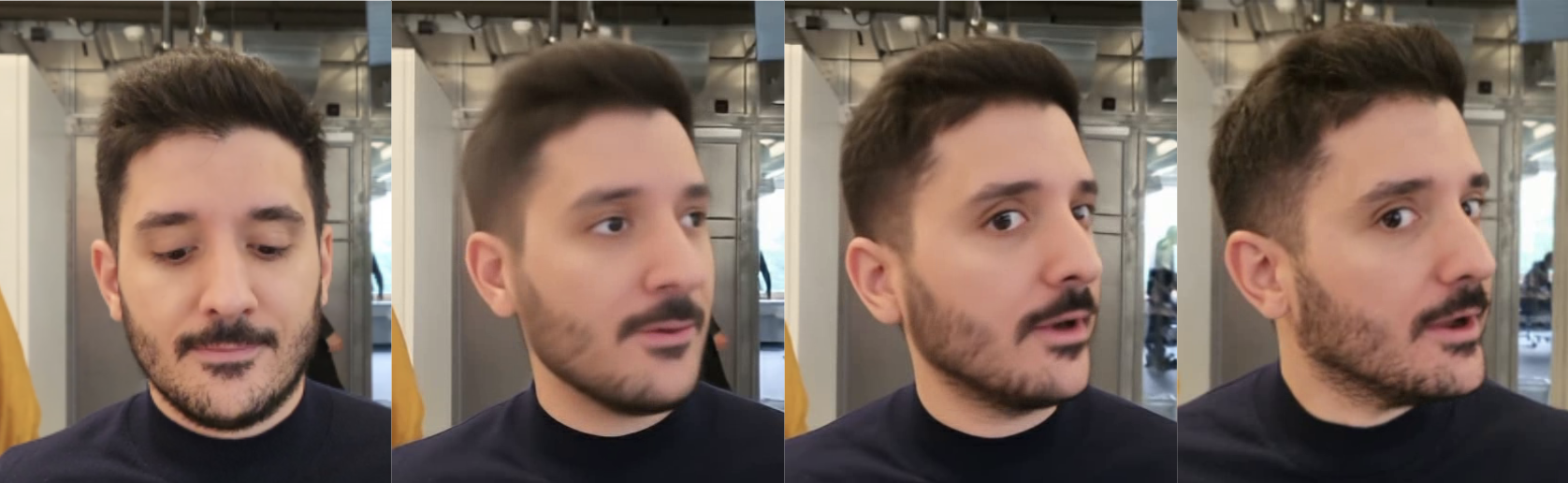}
\put(-220,-8){\scriptsize{Input Image}}
\put(-169,-8){\scriptsize{Not personalized}}
\put(-104,-8){\scriptsize{Personalized}}
\put(-49,-8){\scriptsize{Groundtruth}}
\caption{Qualitative results on model personalization. Finetuning our model~\cite{dreambooth} on a single video of a user supports more veridical synthesis over a wide range of expressions.
\label{fig:results_personalization}}
\end{figure}

\paragraph{\bf Video Editing.} 
Similarly, our diffusion approach exhibits capabilities in video editing. \cref{fig:results_editing} shows editing examples given an input video (top row) by closing the mouth (second row), eyes (third row) or keeping the subject's eyes open, \eg not blinking (third row), in a temporally coherent manner. 
In this case, we automatically generate an inpainting mask based on the body coordinates that project differently than in the groundtruth image, after editing their face expression, and use this temporal mask to re-generate the pixels according to the new target controls. 
This process is independent of the length of the video, distance to camera, or subject identity, and we hope these results can lead to novel applications on creative video editing. See videos in Sup. Mat.

\paragraph{\bf Personalization.} 
Personalization in the context of diffusion models has been extensively explored recently for subject-driven generation~\cite{dreambooth}. In our case, \Model only takes a monocular input image as source for synthesis,
and while it can produce a plausible synthesis, it has no access to occluded parts and the resulting video may not be veridical at a fine grain analysis of that person. %
In \cref{fig:results_personalization}, we show that by fine-tuning our diffusion model with more data, on a monocular video of a subject, \Model can learn to capture the identity better, \eg when the reference image displays the eyes as closed. 

\section{Conclusion}
\label{sec:conclusion}

We have presented \Model, a methodology for human video synthesis, including both face and body, from a single input image, conditioned by audio or text. \Model is built as a temporal extension of control-based diffusion models, with underlying scaffolding based on 3d human head and body pose representations, which generates high quality animations of variable length. We introduce a diverse and large scale dataset (one order of magnitude larger than previous ones), and validate the performance of \Model on this and multiple other repositories, showing that it outperforms previous state-of-the-art on the task of talking face generation, and that our approach is more robust on different diversity axes. Sup. Mat. discusses limitations and societal impact.

\noindent{\bf Acknowledgements}: 
We gratefully acknowledge Alonso Martinez, Anja Hauth, Sergi Caelles,
Hernan Moraldo, Erik Frey, Krishna Somandepalli and Brendan Jou for their careful collection and analysis of a large and diverse repository of videos from which we curated MENTOR.

\bibliographystyle{splncs04}
\bibliography{main}
\end{document}